\documentclass{article}

\usepackage[preprint]{neurips_2023}

\usepackage[utf8]{inputenc} 
\usepackage[T1]{fontenc}    
\usepackage{hyperref}       
\usepackage{url}            
\usepackage{booktabs}       
\usepackage{amsfonts}       
\usepackage{nicefrac}       
\usepackage{microtype}      
\usepackage{amsmath}
\usepackage{amssymb}
\usepackage{enumitem}
\usepackage[]{graphicx}
\usepackage{caption}
\usepackage{subcaption}
\usepackage{wrapfig}     
\usepackage{indentfirst}

\title{Balancing Explainability-Accuracy of Complex Models}

\author{%
 Poushali Sengupta\\
  Department of Informatics\\
  The University of Oslo\\
  Oslo, Norway \\
  \texttt{poushals@uio.no} \\
   \And
   Yan Zhang \\
  Professor \\
  Institute for Informatics\\
  The University of Oslo\\
  Oslo, Norway\\
  \texttt{yanzhang@ifi.uio.no} \\
  \And
  Sabita Maharjan \\
  Professor \\
  Institute for Informatics \\
  The University of Oslo\\
  Oslo, Norway\\
  \texttt{sabita@ifi.uio.no} \\
  \And
  Frank Eliassen \\
  Professor Emeritus\\
  Institute for Informatics\\
  The University of Oslo \\
  Oslo, Norway\\
  \texttt{frank@ifi.uio.no} \\
}

\begin{document}

\maketitle

\begin{abstract}
  Explainability of AI models is an important topic that can have a significant impact in all domains and applications from autonomous driving to healthcare. The existing approaches to explainable AI (XAI) are mainly limited to simple machine learning algorithms, and the research regarding the explainability-accuracy tradeoff is still in its infancy especially when we are concerned about complex machine learning techniques like neural networks and deep learning (DL). In this work, we introduce a new approach for complex models based on the co-relation impact which enhances the explainability considerably while also ensuring the accuracy at a high level. We propose approaches for both scenarios of \textit {independent features} and \textit {dependent features}. In addition, we study the uncertainty associated with features and output. Furthermore, we provide an upper bound of the computation complexity of our proposed approach for the dependent features. The complexity bound depends on the order of logarithmic of the number of observations which provides a reliable result considering the higher dimension of dependent feature space with a smaller number of observations. 
\end{abstract}
\section{Indroduction}
\vspace{-2mm}
\begin{wrapfigure}{r}{0.4\textwidth}
    \includegraphics[ clip = true, width=\linewidth, trim={45 65 10 90}]{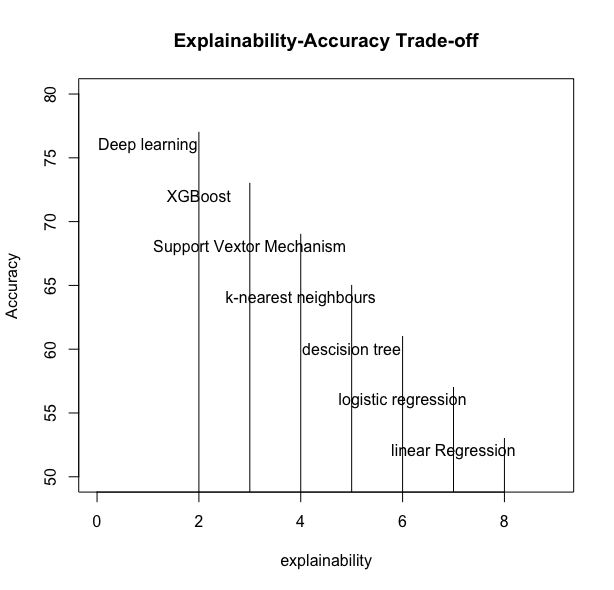}
    \vspace{-2mm}
    \caption{The graph shows the trade-off between explainability-accuracy; More complex models are hard to explain.}
    \label{fig}
\end{wrapfigure}
Artificial Intelligence (AI) is envisioned to play an important role in various technical fields from the energy market to clinical domains[3, 9, 10]. The machine learning algorithm, especially with the recent development in deep learning, has proven to be an indispensable technology to any system including critical infrastructure[12, 13, 14, 15]. As a consequence, there is a growing interest in understanding the execution and performance of these algorithms[22, 23]. Explainable AI (XAI)[1, 2] can be a promising research field to understand and interpret the "black box" behaviour of machine learning (ML) algorithms[24, 25, 26].  

SHAP: Shapley Additive Explanations[18]and LIME[19] are the most used algorithms for the explainability of an ML model. The SHAP-based approach calculates the weighted marginal contribution called Shapley value for each feature. The Shapley value of a feature represents its contribution to one or several sets of features. On the other hand, LIME focuses on the local faithfulness of a  model. Although LIME[19] has the desirable property of additivity[18], it has weaknesses regarding the lack of consistency[16], missingness[17], and stability[20, 27]. SHAP fulfils these and hence is commonly used. The sum of SHAP values associated with all features is equal to the final contribution. LIME assumes that the local model is linear, whereas SHAP can be applied to any simple nonlinear model[21]. However, for complex ML models, the SHAP algorithm simplifies the model first and then calculates the feature importance[28]. SHAP considers all possible combinations of features, which becomes hard to apply with high dimensional dependent feature space. To this end, while more complex ML/DL models yield better accuracy, the explainability of such models is a challenging and largely unexplored problem (see figure \ref{fig} for better understanding). In particular, enhancing the explainability of a model while also ensuring a high level of accuracy is not a trivial task[29]. Also, most of the existing algorithms including SHAP, are hard to apply when features are dependent [30]. Furthermore, existing XAI models including SHAP-based approaches interpret the result without measuring the uncertainty[32, 33] regarding the feature's contributions to the output. On this background, the three main challenges addressed by us,
\textbf{1)}  \textit{The explainability-Accuracy tradeoff in a complex model}, \textbf{2)}  \textit{Large number of dependent input features}, and  \textbf{ 3)} \textit{Uncertainty associated with each feature and output.} \\
Motivated by these considerations, we propose a new approach \textit{ExCIR: Explainability through Correlation Impact Ratio} that can maintain an accuracy-explainability trade-off for complex ML models, considering both dependent and independent features in higher dimensional feature space. ExCIR also considers the uncertainty associated with a feature contribution by measuring Shannon entropy [31] for each feature. ExCIR adjusts the distance between features and output vectors in hyper-dimensional space for both the original and the lightweight model environment to ensure the accuracy of the model.  Instead of SHAPLY values[21, 23], ExCIR calculates the correlation impact ratio ($\mbox{CIR}$) to explain the relation distance of a target feature on the output variable. 
The main contributions of this work are:
\vspace{-3mm}
\begin{enumerate}[noitemsep]
    \item \textbf{Novel framework for Accuracy-Explainability tradeoff:} We create a lightweight model to explain the feature impacts on the output by creating a suitable environment such that the lightweight model works with sample data having less input instead of choosing different sets of features. 
    \item \textbf{Introducing new metric to calculate the uncertainty of explainability:} To calculate this metric we will consider the notion of uncertainty [6] which measures the uncertainty between the features and output.
     \item \textbf{Novel metrics of Explainability for both feature independencies and dependencies:  } We propose different measures to compute the relation impact for the cases of independent features and dependent features. Both the metrics can be directly applied to the nonlinear model and as a result, the explainability is enhanced while ensuring the same level of accuracy. 
\end{enumerate} 

\section{ExCIR: System Model and Design}
\renewcommand{\vec}[1]{{\underset{\sim}{#1}}}
We consider a data set with $n$ rows and $k$ features, denoted as $\vec{f_1},\vec{f_2}, ..., \vec{f_k}; \ \vec{f_i }\epsilon ||F||^{(k\times n)} $ where $||F||^{(k\times n)} $ is the $(k\times n)$ dimensional feature space having $k$ different features with the $n$ observations. Here, the $i^{th}$ feature is denoted as  $\vec{f_i}; \ i= 1:k$ and $\vec{f_i} = (f_{i1}, f_{i2}, ....., f_{in})$. The machine learning model $M(f)$ learns all data history and predicts the output vector $\vec{Y} = (y_1, y_2, ..., y_2)$. We are interested in finding how the features are related to each other and how it affects the result. The correct and detailed explanation can help the model to perform better and reduce the loss in future predictions. We derive a lightweight explanation model that works with the sample of the original data. We create a suitable environment for our XAI model where the accuracy of the lightweight model is almost the same as the original model. We achieve the model output accuracy by equating the projection and the embedding distance[37]. This strategy helps to create the twin environment by securing the same positions of input data distributions in both high-dimensional spaces. As the environments are similar for the original and lightweight models, the explainability of the lightweight model is good enough to explain the original model. The reason for using a lightweight model is that a higher dimension of input data with bigger feature space can make ExCIR more complicated to apply in real-life practice. Also, we calculate mutual information and uncertainty between features and the output to explain the impact of the dependent feature because lower dimensional input space provides lesser bias in computing the uncertainty associated with the features. The main difference between our ExCIR model with the surrogate model[39] is, the ExCIR model is a copy of the original model which works with the same data set and chooses multiple numbers of combinations of features to calculate feature importance. On the other hand, our model is equivalent to the original model but works with less data and considers all features for the calculation of the feature impact ratio. In ExCIR, we do not need to consider different combinations of feature sets, whereas we calculate Shannon entropy considering the uncertainty of the features, and based on that we introduce Conditional Multivariate Mutual Information $\mbox{CMMI}$. The main reason for achieving the same environment is,
\vspace{-3mm}
\begin{enumerate}[noitemsep]
    \item  Both the lightweight model and the original model's environments are equivalent, so we are securing the accuracy guarantee at first. 
    \item We apply our new XAI approach to the lightweight model and as the lightweight model works with a lower-dimension environment, it is more flexible to calculate complex metrics like $\mbox{\mbox{CMMI}}$ for each feature.
    \item As we prove that both models have the same environment (see next section), we can claim the impact of the feature in the lightweight model will be the same as the original one, i.e., the feature's contributions will be the same for both models.
\end{enumerate}
\par Let $M'$ denote the lightweight model, trained over the lower dimensional data set having the same number of features $k$ as $\vec{f_1},\vec{f_2}, ..., \vec{f_{k}}; \ \vec{f_i }\epsilon ||F||^{k \times n'} $ and $\vec{f_i} = (f_{i1}, f_{i2}, ....., f_{in'})$, where $i=1: k$ and $n'; n'< n$ is the number of rows in the data set. Let $Y'$ denote the output variable.\\
 $||F||^{(k\times n)}$ is an $(k \times n)$ dimensional feature space that contains each feature's distributions. A feature distribution refers to a distribution that is followed by all the data points of specific features. So we can consider unified $([k+1] \times n)$ dimensional super space $\mathcal{U}$ such that $||F||^{(k\times n)}  \subset \mathcal{U}$. $\mathcal{U}$ contains all feature distributions as well as target output distribution. We assume that $(k+1)$th distribution is the output distribution. Let, $\mathcal{D}_i; \ i=1:[k+1]$ denote the distribution of the $i ^{th}$ feature, and $\mathcal{D}(Y)$ is the output distribution of the original model. Then we can have, $\mathcal{U} = [\mathcal{D}_1(\vec{f_1} ) \cup  \mathcal{D}_2(\vec{f_2} )  \cup ... \cup \mathcal{D}_{k}(\vec{f_{k}} ) \cup \mathcal{D}(Y) )]  $. On the other hand,  $\mathcal{D}(Y') $ is the output distribution from the lightweight explainable model. Without loss of generality, $\mathcal{U'}$ is the superspace for the lightweight model where $\mathcal{U'} \subset \mathcal{U}$  and we can have, $\mathcal{U'} = [\mathcal{D}_1(\vec{f_1} ) \cup  \mathcal{D}_2(\vec{f_2} )  \cup ... \cup \mathcal{D}_{k}(\vec{f_{k}} ) \cup \mathcal{D}(Y') )]. $
Our main idea is to find the relation among features and how it affects the output. These relations can be non-linear and dependent.  ExCIR will work directly with the non-linear model for dependent as well as independent features. 
\section{Accuracy of ExCIR}
To maintain the lightweight model's accuracy, the environment of the lightweight model must be almost the same as the original model. Because within the same input-output environment both the models should behave in the same manner. Here, the environment refers to the features' distributions and their positions in the superspace. Every feature has some impact on generating the output. Keeping this in mind, we equate the distance between each feature and output distribution for both spaces $\mathcal{U}$ and $\mathcal{U'}$. If the distance of the same feature to the output is the same in both spaces, we can claim that the positions of features in both spaces are similar. So, $\mathcal{U'}$  will create a twin environment of $\mathcal{U}$ with a lower dimension. Once we secure the feature and output distribution position, in the next step, we use projection and embedding distance [37] through f-divergence so that the lightweight model output distribution becomes a mirror image of the original model output. More specifically, we can claim that the lightweight model environment is the same as the original model when
\vspace{-3mm}
\begin{enumerate}[label=(\roman*),noitemsep]
    \item The distances between features and output distributions are the same in both spaces.
    \item  Output distributions in both spaces are mirror images of each other.
\end{enumerate}
Let   $\vec{\mathcal{F}} = (\mathcal{F}_1, \mathcal{F}_2, ...., \mathcal{F}_n)'$ denote  the input column vector, 
where $\mathcal{F}_j ; j = 1:n$ is the $jth $ input containing all $k$ features. That means, $\mathcal{F}_j = [f_{1j}, f_{2j}, ....., f_{kj}]$; $j = 1: n$. The output vector is $Y = (y_1, y_2, ...., y_n)'$. 
Now, each of $y_i$ s is explained by the $k$ features of the $jth$ input vector. that means, $y_j$'s are explained by $\mathcal{F}_j$s. Now if we consider a $k$ dimensional input space, then by the definition of Euclidean distance [38] we define the local distance for a single output $y_i$ as follows: 
\vspace{-3mm}
\begin{equation}
 D^2_{ij} =  \sum_{j= 1}^{k} (y_i - f_{ji})^2
\end{equation}
The average  $k$ dimensional distance between $Y$ and $\vec{\mathcal{F}}$ can provide the exact position of the distribution of $Y$ in a super space. It will help us build our lightweight model environment to produce a highly accurate output space with similar influences of features as the original data. So to calculate the distance between $Y$ and $\vec{\mathcal{F}}$, we have to consider all local distances. The final average distance for the original and sample space can be defined as respectively: 
\vspace{-3mm}
\begin{equation}
    D^2_\text{final} = \frac{1}{n}\sum_{j= 1}^{k} \sum_{i= 1}^{n} (y_i - f_{ji})^2, \ \text{and} \ D'^2_\text{final} = \frac{1}{n'}\sum_{j= 1}^{k} \sum_{i= 1}^{n'} (y'_{i} - f_{ji})^2  
\end{equation}
The idea is to equalize two distances so that the output distribution for two data sets has the same distance from their feature i.e., exact same positions in the high dimensional space. We will have to achieve the same output environment in lightweight space so that we can get the exact accurate relation impact of each feature on the output vector. To create such a situation we have to minimize the difference between two distance measurements which means $|D^2_\text{final} - D'^2_\text{final}| \rightarrow 0$. 
We define the following condition for achieving an equivalent environment for the lightweight explainable model:
\par \textbf{Definition 1:}
\textit{For an ML model, let $\vec{\mathcal{F}}$ be the input vector having $k$ number of features, $Y$ be the output vector and for a lightweight model, $\vec{\mathcal{F}'}$  be the input vector having the same $k$ features and $Y'$ be the output vector. In a multidimensional space, the average distances from the feature space to the output vector for the original ML model and for the lightweight model are denoted as $D^2_\text{final}$, and $D'^2_\text{final}$ respectively. Then we can claim that both the spaces' features and output distribution have similar positions with the exact same amount of relation impact of each feature if and only if : }
\vspace{-1mm}
\begin{equation}
    \lim_{y\rightarrow y'}|D^2_\text{final} - D'^2_\text{final}| =0
\end{equation}

More fundamentally, when two spaces have the equivalent environment and the output distributions are in similar positions i.e., the average distance from one local output to the corresponding feature input is approximately the same for both spaces and given the fact that two distributions are exactly mirror image of each other (proved later in this section), we can claim that the impact of the $ith$ feature on the output vector is exactly the same in both multi-dimensional spaces. So, if we consider the lightweight space to calculate the relation impact of each feature for generating the report,  it will be able to accurately explain the original model without loss of quality in the result. So, in the next step, we will minimize the distance between the original and the lightweight model output distribution in different dimensions to achieve the mirror effect.  \\
In ExCIR, after passing the sample dataset to the lightweight model, the model will generate an initial output, defined as $Y'$. Then we will minimize the loss function $\mathcal{L}(Y, Y')$ with respect to the distance from output distribution to feature distributions. Now the lightweight model super space  $\mathcal{U'}$  and original model super space  $\mathcal{U}$ can be of different dimensions, as we are working with the low dimensional sample data set. In this case, we need to consider calculating the output distribution distance which belongs to a different dimension space that makes it challenging to compute distances. At first, we will use two different approaches Projection distance and Embedded distance [37] to measure the distance of output distributions of the original and lightweight model. 

let, $m(\Omega)$ denotes all sets of borel probability measures where $\Omega \  \epsilon \ \mathbb{R}^n $, and let $m^{p}(\Omega) \subseteq m(\Omega)$ refer those finite $pth$ moment , where $p \ \epsilon \ \mathbb{N}$. Then for any $n', n \  \epsilon \ \mathbb{N}$ and $n' \ \leq \ n$, we can write 
\vspace{-2mm}
\begin{equation}
    O(n',n) = \{ P \epsilon \mathbb{R}^{n'\times n} : PP^T = I_{n'} \}   
\end{equation}
the $n' \times n$ , metrics consists of orthonormal rows and we can write $O(n) = O (n,n)$ for the orthonormal group. $P^T= $ Transpose of $P$ where for any $P \ \epsilon \ O(n',n)$ and $b \ \epsilon \ \mathbb{R}^{n'}$,
\begin{equation}
    \Phi_{P,b} : \mathbb{R}^n \rightarrow \mathbb{R}^{n'},  \Phi_{P, b(x)} = Px + b;   
\end{equation}
and for any $\mu \ \epsilon \ m(\mathbb{R}^n)$, $\Phi_{P, b(\mu)} = \mu \circ \Phi_{P, b}^{-1}$ is the pushforward measure where $\Phi_P = \Phi_{P,0}$ if $b =0$. More specifically, for any measurable map $\Phi : \mathbb{R}^n \rightarrow \mathbb{R}^{n'}$,  $\Phi (\mu) = \mu \circ \Phi$ is the pushforwad measure. 
Here for our method, without loss of generality, we assume that $\mu = \mathcal{D}({Y})$ is the lightweight model output probability measure and $\delta = \mathcal{D}(Y)$ is the original model output probability measure. We define a distance $d(\mu, \delta)$ where, $\mu \ \epsilon \ m(\Omega_1)$ and $\delta \ \epsilon \ m(\Omega_2)$; $\Omega_1 \subseteq \mathbb{R}^{n'} $, and $\Omega_2 \subseteq \mathbb{R}^{n}$. Here, $n, n' \ \epsilon \ \mathbb{N} $ and $n' \leq n$. We consider here the notion of f-divergence to measure distance function. According to [37] any method of f-divergence (kl divergence [42], Jensen- Shanon divergence [41] etc ) will satisfy the algorithm. Without loss of generality, we can consider,
\vspace{-2mm}
\begin{equation}
    \Omega_1 = \mathbb{R}^{n'} ,  \Omega_2 = \mathbb{R}^{n}
\end{equation}
Now we are interested in calculating the distance with projection and embedding measures [37]. \\
\textbf{Definition 2: } \label{def 1}
let $n',n \ \epsilon \ \mathbb{N}$, $n' \leq n$. For any $\mu \ \epsilon \ m(\mathbb{R}^{n'})$ and $\delta \ \epsilon \ m(\mathbb{R}^n)$, the embedding of $\mu$ into $\mathbb{R}^n$ are the set of $n$-dimensional measures 
\begin{equation}
    d^+ (\mu, n) = \{\alpha \ \epsilon m(\mathbb{R}^n) : \Phi_{P,b}(\alpha) = \mu \ \text{for  some} \  P \epsilon O(n',n), b  \ \epsilon \ \mathbb{R}^{n'}\};
\end{equation}
and, \\
\textbf{Defination 3:} \label{def 2}
let $n',n \ \epsilon \ \mathbb{N}$, $n' \leq n$. For any $\mu \epsilon \  m(\mathbb{R}^{n'})$ and $\delta \epsilon m(\mathbb{R}^n)$. The projection of $\delta$ onto $\mathbb{R^{n'}}$ are the $n'$- dimensional measures,
\def\Inf{\operatornamewithlimits{inf\vphantom{p}}}
\vspace{-1mm}
\begin{equation}
\vspace{-1mm}
    d^- (\delta, n') = \{ \beta \ \epsilon \ m(\mathbb{R}^{n'}) : \Phi_{P,b}(\beta) = \delta \ \text{for some } \ P \ \epsilon \ O(n',n), b \  \epsilon \ \mathbb{R}^{n'}\}
\end{equation}
Let $d$ be any notion of distance on $m(\mathbb{R}^{n})$ for any $n \ \epsilon \ \mathbb{N}$.  Then the projection distance will be :
\begin{equation}
\vspace{-2mm}
    d^-(\mu, \delta) = \Inf_{\beta \in d^+(\delta, n')} \text{d}(\mu, \beta)
\end{equation}
and the embedding distance will be,
\begin{equation}
    d^+(\mu, \delta) = \Inf_{\alpha \in d^+(\mu, n)}\text{d}(\delta,\alpha)
\end{equation}
Both $d^-(\mu, \delta)$ and $ d^+(\mu, \delta) $ calculate the distance between two probability measures $\mu$ and $\delta$  of different dimensions. In [37] it is shown that if $d$ is an f-divergence, then  $d^-(\mu, \delta) = d^+(\mu, \delta)=\hat{d} (\mu, \delta) $. The authors in [37] also generalized the theorem and proved that $d^-(\mu, \delta) = d^+(\mu, \delta)=\hat{d} (\mu, \delta) = 0 $ if and only if $\Phi_{P,b}(\delta) = \mu $ for
some $P \ \epsilon \ O(n',n)$, and $b \ \epsilon \ \mathbb{R}^{n'}$. This means if the expected distance $\hat{d} (\mu, \delta)$ can be minimized to $0$, then the necessary and sufficient condition for claiming that two probability measures  $\mu$ and $\delta$ are rotated and translated copies of each
other, modulo embedding in a higher-dimensional ambient
space where $n' \neq n$. 
Here, our loss function is the expected distance and the goal is to minimize the expected distance between the two output distributions $\mu = \mathcal{D}(Y')$ and $\delta = \mathcal{D}(Y)$ as close as $0$ through a risk generator. This will help to boost the accuracy of the model. 
So,  we can define the loss function as :
\vspace{-2mm}
\begin{equation}
    \mathcal{L}(Y, Y') = E_{\mu \epsilon m(\mathbb{R}^n), \delta \epsilon m(\mathbb{R}^{n'}) }(\hat{d} (\mu, \delta)) = E_{Y \epsilon \mathcal{U}, Y' \epsilon \mathcal{U'} }(\hat{d} (\mathcal{D}(Y), \mathcal{D}(Y')))
\end{equation}
\vspace{-2mm}
\begin{equation}
    \mathcal{L}(Y, Y') = E[\hat{d} (\mathcal{D}(Y), \mathcal{D}(Y'))| Y \epsilon \mathcal{U}, Y' \epsilon \mathcal{U'} ]
\end{equation}
To minimize the loss function, we introduce a risk generator function, say $\mathcal{R}(d*) \epsilon \mathcal{H}$ which will give us the desired result for higher accuracy. Here, $\mathcal{H}$ is a hypothesis class that contains all possible risk generators. We will choose the final risk generator $\mathcal{R}(d*)$ when $\hat{d}(\mathcal{D}(Y), \mathcal{D}(Y')) \rightarrow 0$.
We can define $\mathcal{R}(d*)$ as: 
\def\argmin{\operatornamewithlimits{argmin\vphantom{p}}}
\vspace{-3mm}
\begin{equation}
    \mathcal{R}(d*) = \argmin_{\mathcal{R}(d) in \mathcal{H},\hat{d}\rightarrow 0 } \int_{\mathcal{D}(Y) \epsilon \mathcal{U}} \int_{\mathcal{D}(Y') \epsilon \mathcal{U'}} E [\mathcal{L}(Y, Y')] + \lambda(.)
\end{equation}
Here $\lambda$ is the regularized parameter. Clearly from the equation, we can see when $E[\mathcal{L}(Y, Y')] = \mathcal{L}(Y, Y') \rightarrow 0$ , then $\hat{d} \rightarrow 0$ implies $Y \rightarrow Y'$. So, it will achieve the necessary and sufficient condition where two probability measures  $\mathcal{D}(Y)$ and $\mathcal{D}(Y')$ is rotated and translated copies of each other, and  $n' \neq n$. As a result, the accuracy of the output is the same as the original model.
\section{Explainability of ExCIR}
\begin{wrapfigure}{l}{0.4\textwidth}
    \includegraphics[width=\linewidth, height=22mm]{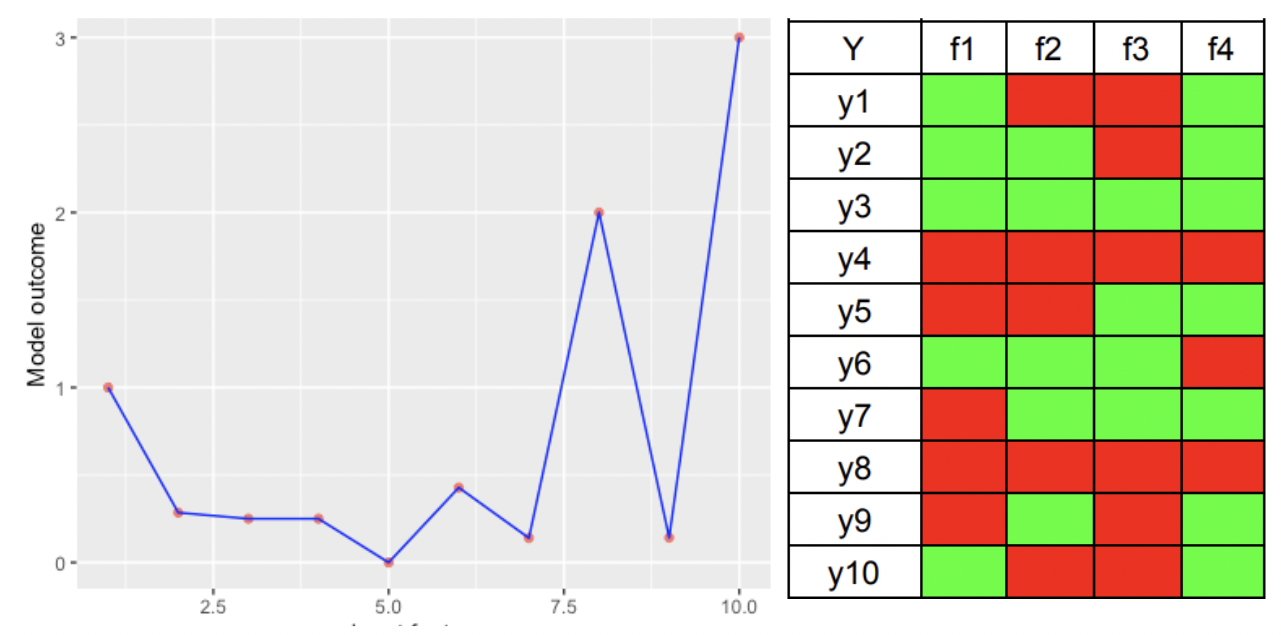}
    \caption{The graph reflects the behaviour of the model in equation \ref{nonlinear} considering the feature's uncertainty to contribute to output generation and the table indicates the presence and absence of the feature for generating output, green represents the presence, and red represents absence.}
    \label{fig_2}
\end{wrapfigure}
We use $\mbox{CIR}$ to measure the feature relation impact on the final output and show how it works in two different environments regarding \textit{independent} and \textit{dependent} features.
 To check the relation between a feature $\vec{f_1}, \vec{f_2}, ..., \vec{f_k}$ and output (linear or nonlinear) we need to hold one parameter constant and also hold input features $(\vec{f_1})$ constant. Then we vary the remaining parameters and watch how $\vec{Y'}$ changes. If the change in $\vec{Y'}$ is non-linear with the change in the parameter which is varied, and if this is true for all the parameters in the model, then that model is said to be non-linear."Non-Linear" describes the model, not the graph of $(\vec{f_1}...\vec{f_k})$ vs. $\vec{Y'}$. 

An example of non-linear regression would be something like this: 
 \vspace{-1.5mm}
 \begin{equation} \label{nonlinear}
M(f_{(f,\beta)})= \frac{\beta_1 f_1 + \beta_2 f_2}{\beta_3 f_3 + \beta_4 f_4 + ....... + \beta_n f_n }
 \end{equation}
The equation \ref{nonlinear}, may behave like a linear model depending on parameter choosing in the general case. In our work, as we consider the uncertainty associated with the feature's contribution i.e., a feature may or may not contribute to generating an output, equation \ref{nonlinear} should behave like a nonlinear model. For a set of empirical outputs, figure \ref{fig_2} demonstrates the curve of how the model in equation \ref{nonlinear} would behave considering the feature's uncertainty. For our model, we also assume that at least two features (one is conversely and another is directly related to the output) are present.


 \subsection{Partial correlation Impact ratio for independent features}\label{ind}

We have a data set with $n'$ rows and $k$ features, denoted as $\vec{f_1},\vec{f_2}, ..., \vec{f_k} ; \vec{f_i} \epsilon ||F||^{k \times n'} $. The machine learning model $M'(f)$ learns all data history and predicts the output vector $\vec{Y}' = (y'_1, y'_2, ..., y'_{n'})$. Changing the notation for more easy understanding and without loss of generality,
let $f_{ij}$ denote the $jth$ observation of the $ith$ feature; where $i= 1:k$ and $j= 1:n'$ and let $y'_j$ denote the {jth} observation of the output vector $\vec{Y'}$; $j= 1:n'$.
then the mean of the feature $f_i ; i = 1:k$, the mean of the output vector $Y$ will be, and the joint weighted mean of feature $f_i$ ; $i = 1:k$ and $Y$ will be respectively,
\vspace{-3mm}
\begin{equation}
    \hat{f_i} = \frac{\sum_j f_{ij} }{n'}, \  \hat{y'} = \frac{\sum_j y_j}{n'},  \  \text{and} \ \hat{f_iy'} = \frac{\hat{f_i} + \hat{y'}}{2}
\end{equation}
To measure the impact of each feature on the output, we calculate the relation impact for each feature separately. Then we define the $\mbox{PCIR}$ of feature $\vec{f_i}$ on the output vector $\vec{Y'}$ by a ratio $\eta_{f_i} = {\beta_{f_i}}^2$ which is formulated as:
\begin{equation}
    \eta_{f_i} =   \frac{n'[(\hat{f_i}-\hat{f_{i}y'})^2 + (\hat{y'}-\hat{f_{i}y'})^2]}{\sum_j (f_{ij} - \hat{f_{i}y'})^2 + \sum_j ( y'_j - \hat{f_{i}y'})^2 }
\end{equation}

i.e., the weighted variance of the $ith$ feature and output variable with respect to joint mean $\hat{f_{i}y'}$ divided by the variance of all values with respect to joint mean $\hat{f_{i}y'}$.
  This correlation impact ratio $\eta_{f_i}$; $i = 1 : k$ takes values between $0$ and $1$. 

For $\eta_{f_i} = 1$, the dispersion is the same for both the output and features. That means a small change in feature observation leads to changes in output observation. So, we can say the $ith$ feature has a great impact on output. Whereas, $\eta_{f_i} = 0$ refers to the case when $ith$ features have almost no impact on the output. We divide the features into two groups, the first group is directly related to output; i.e., changes in the same direction. On the other hand, another group of features is conversely related to the output; i.e., (changes in a different direction). For the sake of simplicity, we take $(\vec{f_1},\vec{f_2},....,\vec{f_m}); 0<m \leq k$ directly related to output and belongs to numerator and $\vec{(f_{(m+1)}},\vec{f_{(m+2)}},....,\vec{f_k)}$ conversely related to output and belongs to denominator.    
Then we can write ExCIR model as: 
\vspace{-1mm}
\begin{equation}
 M'(f_{(f,\beta^2)})= \frac{\beta_{f_1}^2 \vec{f_1} + \beta_{f_2}^2 \vec{f_2} + ...... + \beta_{f_{m}}^2 \vec{f_{m}}}{\beta_{f_{m+1}}^2 \vec{f_{m+1}} + \beta_{f_{m+2}}^2 \vec{f_{m+2}} + ....... + \beta_{f_{k}}^2 \vec{f_{k}} }
 \end{equation}
 where $0< m \leq k$ and we can modify it to the following form : 
 \vspace{-2mm}
 \begin{equation}
    M'(f_{(f,\eta)})= \frac{\eta_{f_1} \vec{f_1} + \eta_{f_2} \vec{f_2} + ...... + \eta_{f_{m}} \vec{f_{m}}}{\eta_{f_{m+1}} \vec{f_{m+1}} + \eta_{f_{m+2}} \vec{f_{m+2}} + ....... + \eta_{f_{k}} \vec{f_{k}} }
 \end{equation}
 For the $jth; j =1:n $ observations of features $f_{1j},f_{2j}, ....., f_{kj} $ , the local output $y_j $ can be expressed as :
 \vspace{-2mm}
 \begin{equation} \label{local}
   y'_j =  M'(f_{j(f,\eta)})= \frac{\eta_{f_1} f_{1j} + \eta_{f_2} f_{2j} + ...... + \eta_{f_{m}} f_{mj}}{\eta_{f_{m+1}} f_{m+1,j} + \eta_{f_{m+2}} f_{m+2,j} + ....... + \eta_{f_{k}} f_{kj} }
 \end{equation}
 
 Here , $f_{ij} ; i=1:k, j=1:n$ is the binary variable. $f_{ij}=1$, refers to the presence of the $ith$ feature in the $jth$ entry. If we can prove that the changes in the output by changing a single input in the $ith$ feature are actually dependent on its corresponding correlation ratio, i.e., $\eta_{f_{i}}$, we can claim that our proposed correlation ratio can reassure the feature importance.

 \textbf{Theorem 1:}\label{them1} \textit{The changes in a particular feature input can affect the changes in local output. So the changes must either depend on the corresponding correlation ratio of that feature or is constant, given that all the features are independent of each other. that is: }
 \vspace{-2mm}
 \begin{equation}
 \begin{aligned}
     &\frac{dy}{d\vec{f_j}} = c_1 . \eta_{f_j} ; \text{ when $j$ < $m$}, \text{, and \ } \frac{dy}{d\vec{f_j}} =\frac{c_2}{(2 K_2-\eta_{f_j}^2) } ;  \text{when $j$ $\geq$ $m$}
\end{aligned}
 \end{equation}
 Here, $c_1, c_2,$
 and $k_2 $ are constants. proof is in supplementary subsection \ref{thm1}.

\textbf{Corollary 1: } \textit{When the particular feature is positively related to the output, the expected change in the output due to the change in the input feature is directly proportional to its correlation ratio impact. On the other hand, if the particular feature is negatively related to the output, the expected change in output due to the change in input of that feature is inversely proportional to its correlation ratio impact. We can express it mathematically as follows:  }\\
\vspace{-2mm}
 $E(\frac{dy'}{d\vec{f_j}}) \varpropto \eta_{f_j}; $if $j \leq m$ , and 
$E(\frac{dy'}{d\vec{f_j}}) \varpropto \frac{1}{\eta_{f_j}}$ if $j > m$ 
\subsection{Mutual Correlation Impact Ratio for dependent features}
In section \ref{ind}, we introduced the correlation impact ratio for the nonlinear model assuming that the features are independent of each other. Though the proposed $\mbox{PCIR}$ is easy to compute and able to keep the balance between interpretability and accuracy, $\mbox{PCIR}$ is only suitable for the environment where the features are independent. So, the current section is only dedicated to the interpretability theorem when the features are dependent. For the multivariate dependent feature space, it is challenging to compute the relation impact of a targeted feature on output due to the influence of other features. To address the issue, we propose a new metric called Mutual Correlation Impact Ratio (MCIR). The state-of-the Conditional Mutual Information $\mbox{CMI}$ [45] considers the situation where more than one feature depends on the same feature. But, in our case, every feature is dependent on multiple features and for this reason, we can not use the existing \mbox{CMI} concept. We also propose a new concept called \textit{Conditional Multivariate Mutual Information} ($\mbox{CMMI}$) considering the case when every feature is dependent on other features. $\mbox{MCIR}$ is calculated based on ($\mbox{CMMI}$) assuming that the given features $(\vec{f_1}, \vec{f_2}, ..., \vec{f_k} \epsilon ||F||^{(k \times n')})$ follow either multivariate probability density function (pdf) or multivariate probability mass function(pmf); features can be both discrete and continuous. 
Our proposed metric $\mbox{CMMI}$ represents the mutual dependency among the targeted feature and the output variable considering the targeted feature is dependent on the rest of the features belonging to a multidimensional feature space. When two non-identical conditional probability distributions are taken, there should be a divergence between their cross-entropy and their individual entropies [43]. In a multivariate environment, this divergence can be called Joint Mutual Information ($\mbox{JMI}$). $\mbox{MCIR}$ is calculated based on ($\mbox{JMI}$), and $\mbox{CMMI}$. $\mbox{MCIR}$ measures how the changes in the targeted feature's input affect the output of the model. It also considers the measurement of uncertainty associated with the feature's contributions. 
At first, for the sake of simplicity, we consider $\vec{f_i}$ depends on  $\vec{f_j}$, while $\vec{f_i}$ is independent of the rest of the features; $i = 1(1)k, j = 1(1)k$, and $i \neq j$. Then, we find the impact of $\vec{f_i}$ on $\vec{Y'}$, given the fact that $\vec{f_i}$ depends on  $\vec{f_j}$. This impact can be explained by the information theory, if and only if we can compute $I(\vec{Y'}; \vec{f_i} |\vec{f_j}); \forall i,j = 1(1)k, i\neq j$. The previously described MI can not provide the desired result. To achieve our goal we have to first calculate the Conditional Mutual Impact [44] [45].

The  conditional mutual information between the output variable $\vec{Y'}$ and the target feature $(\vec{f_i}|\vec{f_j})$, $\forall i,j = 1(1)k;  i\neq j$ is defined as [44]:
\begin{equation}
\begin{aligned}
    &I(\vec{Y'}; \vec{f_i}|\vec{f_j}) = I(\vec{f_i}, \vec{f_j})- I(\vec{Y'}, \vec{f_i}|\vec{f_j}) \\
                &=  \sum_{[f_j=\mathrm{f};f_j\epsilon ||F||^{k \times n'}]} \sum_{[f_i=\mathrm{f^*}; f_j\epsilon ||F||^{k \times n'}] }
    \sum_{Y'=y; Y'\epsilon ||Y'||^{n'}}P(\mathrm{f^*},\mathrm{f}, y ) \log_2[\frac{P(y, \mathrm{f^*}| \mathrm{f})}{P(y | \mathrm{f}) P(\mathrm{f^*} |\mathrm{f})} ]
\end{aligned}
\end{equation}
If any of the features $f_i, f_j$, or $Y'$ is continuous, the summation operator can be replaced by the integral operator. 

\subsubsection{ MCIR; with two dependent features} 

When any two of the $k$ features are dependent on each other and other features are independent, the state-of-the-art $\mbox{CMI}$ is sufficient to explain the mutual dependency of $Y'$ and $(f_i |f_j)$. But the value of $\mbox{CMI}$ varies from $0$ to $\infty$ which is an open bound, thus making scalability a major challenge. So, to scale it down in between $[0,1]$, we derive $\mbox{MCIR}$ and as,
\begin{equation}\label{mcir}
\begin{aligned}
     C(\vec{Y'}; \vec{f_i}|\vec{f_j}) &= \frac{I(\vec{Y'}; \vec{f_i}|\vec{f_j})}{I(\vec{Y'}; \vec{f_i}|\vec{f_j})+I(\vec{Y'},\vec{f_i},\vec{f_j})} 
\end{aligned}
\end{equation}
where, $0\leq I(\vec{Y'}; \vec{f_i}|\vec{f_j})\leq \infty$ and $0\leq I(\vec{Y'},\vec{f_i},\vec{f_j})\leq \infty$. Then, $0\leq\frac{I(\vec{Y'}; \vec{f_i}|\vec{f_j})}{I(\vec{Y'}; \vec{f_i}|\vec{f_j})+I(\vec{Y'},\vec{f_i},\vec{f_j})}\leq 1$. So we can claim, $0\leq  C(\vec{Y'}; \vec{f_i}|\vec{f_j}) \leq 1$.
 $C(\vec{Y'}; \vec{f_i}|\vec{f_j})$ is considered as $\mbox{CMI}$ of the target feature $\vec{f_i}$ on output variable $\vec{Y'}$ when $\vec{f_i}$ depends on $\vec{f_j}$.  $C(\vec{Y'}; \vec{f_i}|\vec{f_j})$ can capture how the changes in inputs of target variable $\vec{f_i}$ can affect the output variable $\vec{Y'}$ considering the fact that the input of $\vec{f_j}$ may or may not be changed. $I(\vec{Y'},\vec{f_i},\vec{f_j})$ is called Joint Mutual Information $(\mbox{JMI})$. For simplifying the notations, we consider a simple environment where  $\vec{f_1}$ and $\vec{f_2}$ are dependent on each other and $\vec{f_3}, \vec{f_4}, ...., \vec{f_k}$ are independent features. We also assume that, $(\vec{f_1}, \vec{f_3}, ...., \vec{f_m})$ are directly related to output variable $\vec{Y}$ and $(\vec{f_2}, \vec{f_{p}}, ....\vec{f_k})$ are conversely related to $\vec{Y'}$; $m, p \leq k$. Then using \ref{mcir} on equation (\ref{nonlinear}), our proposed ExCIR model takes the form: 
\begin{equation}
     E(\vec{Y'})=  M'(f_{j(f,C)}))= \frac{C(\vec{Y'};\vec{f_1}|\vec{f_2}) f_1 + \eta_{f_3} f_3 + ...... + \eta_{f_{m}} f_{m}}{C(\vec{Y'}; \vec{f_2}|\vec{f_1}) f_{2} + \eta_{f_{p}} f_{p} + ....... + \eta_{f_{k}} f_{k} }
\end{equation}
where $\eta_{f_i}; i = 3,...,m,...., p,...,k; m,p\leq k$ is the correlation impact for the independent feature $\vec{f_i}, i = 3,...,m, ..., p ,...,k; m,p\leq k$. $I(\vec{Y'}; \vec{f_1}|\vec{f_2})$ and $I(\vec{Y'}; \vec{f_2}|\vec{f_1})$ can be called $\mbox{MCIR}$s for the dependent features $f_1$ and $f_2$ respectively. But, this theory is only based on the fact that any two of the features are dependent on each other. However, in real-life cases, a feature can be dependent on multiple other features. In that case, one has to consider multivariate distribution considering multiple cases of dependency. Keeping this in mind, we propose a new concept for $\mbox{CMMI}$ in the next section.
\subsubsection{Mutual Correlation Impact Ratio; when all features are dependent}
In ExCIR, it is necessary to calculate the mutual impact of a feature on the output variable while assuming the target feature is dependent on other features. But before we define $\mbox{MCIR}$ for multivariate cases, we have to derive $\mbox{CMMI}$. In [45] the authors derived $\mbox{CMI}$ for a multivariate environment where the features are dependent on another variable. i.e., they considered the case when all variables are dependent on one common variable. But in our work, we address the case when all the features are dependent on each other. So, if we want to calculate the mutual dependence between a targeted feature and the output variable we have to calculate the $\mbox{CMMI}$ given that target feature is dependent on multiple features. More specifically, in existing works [44][45][46], the notion of $I(\vec{Y'}; \vec{f_1}, \vec{f_2}, ..., \vec{f_{k-1}}| \vec{f_k)}$ is derived and used in many real-life cases. However, this approach cannot be directly applied in our environment. We therefore introduce a new matrix $I(\vec{Y'};\vec{f_i} | \{\vec{f_j}\} \subseteq \{||F||^{k \times n'} - \vec{f_i}\}; i\neq j )$]. \\
\textbf{Definition 4: CMMI: } \label{def4}
\textit{If an environment has dataset including $k$ number of dependent features $\vec{f_1}, \vec{f_2}, ..., \vec{f_k} \epsilon ||F||^{k \times n'}$; $f_i = (f_{i1}, f_{i2}, ..., f_{in})$, $i=1(1)k$ with the output variable $\vec{Y'} =(y'_1, y'_2, ....y'_{n'})$, then the $\mbox{CMMI}$ between output variable $Y$ and any targeted feature $f_i$ will be: }
\begin{equation}\label{cmmi}
\begin{aligned}
    & I(\vec{Y'};\vec{f_i} | \{\vec{f_j}\} \subseteq \{||F||^{k \times n'} - f_i\}; i\neq j )) \\
    &=  \sum_{\phi} \int_{\psi}\int_{f_i} \sum_{Y'}P(Y', f_1,.., f_i,..f_k) \log_2[\frac{P(Y'| \{f_j\} \subseteq ||F||^{k \times n'} )}{P(Y'|\{\vec{f_j}\} \subseteq \{||F||^{k \times n'} - f_i\}; i\neq j)}]
\end{aligned}
\end{equation}
where $\phi = f_j \subseteq \{||F||^{k \times n'} - f_i\}_d$ is the set of discrete features, and  $\psi= f_j \subseteq \{||F||^{k \times n'} - f_i\}_c$ - the set of continuous features not including $f_i$. The derivation of the equation \ref{cmmi} is based on the concept of \mbox{MI}[46] and it can be found in the supplementary subsection \ref{thm2}.
 $\mbox{CMMI}$ could take the values between $0$ to $\infty$, unlike the strict bound of $[0,1]$ that the normal correlation coefficient has. 

 \textbf{Definition 5: } \textbf{Mutual Correlation Impact Ratio: } Suppose an environment with a dataset has $k$ number of features $\vec{f_1}, \vec{f_2}, ..., \vec{f_k} \epsilon ||F||^{k \times n'}$ ; $\vec{f_i} = (f_{i1}, f_{i2}, ..., f_{in'})$, $i=1(1)k$ and the features are dependent on each other. The output variable is $\vec{Y'} =(y'_1, y'_2, ....y'_{n'})$. Then, $\mbox{MCIR}$ can be defined as:
 \begin{equation}
\begin{aligned}
   & C(\vec{Y'};\vec{f_i} | \{\vec{f_j}\} \subseteq \{||F||^{k \times n'} - \vec{f_i}\}; i\neq j ))\\ =& \frac{I(\vec{Y'};\vec{f_i} | \{\vec{f_j}\} \subseteq \{||F||^{k \times n'} - \vec{f_i}\}; i\neq j )}{I(\vec{Y'};\vec{f_i} | \{\vec{f_j}\} \subseteq \{||F||^{k \times n'} - \vec{f_i}\}; i\neq j )+ I(\vec{Y'}, \vec{f_1}, \vec{f_2}, ...\vec{f_{j-1}}, \vec{f_j}, \vec{f_{j+1}}, .., \vec{f_k)}}
\end{aligned}
\end{equation}\\
\textbf{Result 1:} $C(\vec{Y'};\vec{f_i} | \{\vec{f_j}\} \subseteq \{||F||^{k \times n'} - \vec{f_i}\}; i\neq j )) $ lies between $0$ and $1$. (See supplementary subsection \ref{rslt1} for the proof).
Then the ExCIR model with $k$ dependent features can be written as :
\begin{equation}
    E(\vec{Y'})=  M'(f_{j(f,\mathfrak{C})})= \mathfrak{J}+\frac{\mathfrak{C}_{f_1} f_1 + ...... + \mathfrak{C}_{f_m}f_{m}}{\mathfrak{C}_{f_p}f_{p} + ....... + \mathfrak{C}_{f_k} f_{k} }
\end{equation}
where,$\mathfrak{C}_{f_i} = C(Y';f_i | \{f_j\} \subseteq \{||F||^n - f_i\}; i\neq j ))$ and $\mathfrak{J} = C(\vec{Y'},\vec{f_1},\vec{f_2}, ..., \vec{f_k})$. $C(\vec{Y'},\vec{f_1},\vec{f_2},...,\vec{f_k}) = \frac{I(\vec{Y'},\vec{f_1},\vec{f_2}, ....,\vec{f_k} )}{I(\vec{Y'};\vec{f_i} | \{\vec{f_j}\} \subseteq \{||F||^{k \times n'} - \vec{f_i}\}; i\neq j )+ I(\vec{Y'}, \vec{f_1}, \vec{f_2}, ...\vec{f_{j-1}}, \vec{f_j}, \vec{f_{j+1}}, .., \vec{f_k})} $ is called Joint Mutual Impact. \\
\par Feature dependencies create a complex environment where computational complexity can not be ignored. By the theorem of Kolmogorov Complexity [48][49], we calculate the computational complexity of ExCIR to generate output $Y'$ when features are dependent and we found that the complexity is upper bounded by $ K_\mathbb{A}(\vec{Y'}| I(\vec{Y'})) + \frac{k}{2}O(\log{n'})+ Const_\mathbb{A}$, where  $\mathbb{A}$ is any other computer and there exist a constant $Const_\mathbb{A}$ for all strings $\vec{Y'} = (y'_1, y'_2, ..., y'_{n'}) \epsilon \{0,1\}$, $I(\vec{Y'}) =$ length of the string of $Y'$. $Const_\mathbb{A}$ does not depends on the output $\vec{Y'}$ [47]. We also found that ExCIR would take no more than $\frac{1}{2}O(\log{n'}) + \zeta_\mathbb{A}$ time complexity to calculate each $\mbox{MCIR}$ associated with features. Here, $\zeta_\mathbb{A}$ is a constant associated with the $ith$ $\mbox{MCIR}$ for any other computer $\mathbb{A}$; $i= 1:k$ (See supplementary subsection \ref{complex} for detailed calculation). Both the complexity bounds depend on the number of observations, not the number of features.  As our algorithm is based on the lightweight model and the less number of data observations, both the upper bounds are very promising.
\section{Conclusion}
ExCIR can balance the tradeoff between explainability and accuracy irrespective of a large number of dependent and independent features. This approach also considers the uncertainty associated with features and output. We provide a time complexity upper bound when features are dependent. The upper bound only depends on input observations and as we used a lightweight model with a sample dataset for our whole approach, the time complexity can be considered as a reliable result.  

\newpage
\section*{References}


\small

[1] Buchanan, Bruce G., and Edward H. Shortliffe. Rule based expert systems: the mycin experiments of the stanford heuristic programming project (the Addison-Wesley series in artificial intelligence). Addison-Wesley Longman Publishing Co., Inc., 1984.

[2] Wick, Michael R., and William B. Thompson. "Reconstructive expert system explanation." Artificial Intelligence 54.1-2 (1992): 33-70.

[3] Guidotti, Riccardo, et al. "A survey of methods for explaining black box models." ACM computing surveys (CSUR) 51.5 (2018): 1-42.

[4] Gunning, David. "Explainable artificial intelligence (xai)." Defense advanced research projects agency (DARPA), nd Web 2.2 (2017): 1.

[5] Nunes, Ingrid, and Dietmar Jannach. "A systematic review and taxonomy of explanations in decision support and recommender systems." User Modeling and User-Adapted Interaction 27 (2017): 393-444.

[6] Holzinger, Andreas. "Interactive machine learning for health informatics: when do we need the human-in-the-loop?." Brain Informatics 3.2 (2016): 119-131.

[7] Roque, Antonio, and Suresh K. Damodaran. "Explainable AI for Security of Human-Interactive Robots." International Journal of Human–Computer Interaction 38.18-20 (2022): 1789-1807.

[8] Zanzotto, Fabio Massimo. "Human-in-the-loop artificial intelligence." Journal of Artificial Intelligence Research 64 (2019): 243-252.

[9] Holzinger, Andreas, et al. "Explainable AI methods-a brief overview." xxAI-Beyond Explainable AI: International Workshop, Held in Conjunction with ICML 2020, July 18, 2020, Vienna, Austria, Revised and Extended Papers. Cham: Springer International Publishing, 2022.

[10] Dwivedi, Rudresh, et al. "Explainable AI (XAI): Core ideas, techniques, and solutions." ACM Computing Surveys 55.9 (2023): 1-33.

[11] Hind, Michael. "Explaining explainable AI." XRDS: Crossroads, The ACM Magazine for Students 25.3 (2019): 16-19.

[12] Atakishiyev, Shahin, et al. "Explainable artificial intelligence for autonomous driving: a comprehensive overview and field guide for future research directions." arXiv preprint arXiv:2112.11561 (2021).

[13] Ohana, Jean Jacques, et al. "Explainable AI (XAI) models applied to the multi-agent environment of financial markets." Explainable and Transparent AI and Multi-Agent Systems: Third International Workshop, EXTRAAMAS 2021, Virtual Event, May 3–7, 2021, Revised Selected Papers 3. Springer International Publishing, 2021.

[14] Tjoa, Erico, and Cuntai Guan. "A survey on explainable artificial intelligence (xai): Toward medical xai." IEEE transactions on neural networks and learning systems 32.11 (2020): 4793-4813.

[15] Machlev, R., et al. "Explainable Artificial Intelligence (XAI) techniques for energy and power systems: Review, challenges and opportunities." Energy and AI (2022): 100169.

[16] Ji, Yingchao. "Explainable AI methods for credit card fraud detection: Evaluation of LIME and SHAP through a User Study." (2021).

[17] Dieber, Jürgen, and Sabrina Kirrane. "Why model why? Assessing the strengths and limitations of LIME." arXiv preprint arXiv:2012.00093 (2020).

[18] Lundberg, Scott M., and Su-In Lee. "A unified approach to interpreting model predictions." Advances in neural information processing systems 30 (2017).

[19] Ribeiro, Marco Tulio, Sameer Singh, and Carlos Guestrin. "" Why should i trust you?" Explaining the predictions of any classifier." Proceedings of the 22nd ACM SIGKDD international conference on knowledge discovery and data mining. 2016.

[20] Visani, Giorgio, Enrico Bagli, and Federico Chesani. "OptiLIME: Optimized LIME explanations for diagnostic computer algorithms." arXiv preprint arXiv:2006.05714 (2020).

[21] Winter, Eyal. "The shapley value." Handbook of game theory with economic applications 3 (2002): 2025-2054.

[22] Goldstein, Alex, et al. "Peeking inside the black box: Visualizing statistical learning with plots of individual conditional expectation." journal of Computational and Graphical Statistics 24.1 (2015): 44-65.

[23] Molnar, Christoph. Interpretable machine learning. Lulu. com, 2020.

[24] Galkin, Fedor, et al. "Human microbiome aging clocks based on deep learning and tandem of permutation feature importance and accumulated local effects." BioRxiv (2018): 507780.

[25] Mehdiyev, Nijat, and Peter Fettke. "Prescriptive process analytics with deep learning and explainable artificial intelligence." (2020).

[26] Ryo, Masahiro, et al. "Explainable artificial intelligence enhances the ecological interpretability of black‐box species distribution models." Ecography 44.2 (2021): 199-205.

[27] Khoda Bakhshi, Arash, and Mohamed M. Ahmed. "Utilizing black-box visualization tools to interpret non-parametric real-time risk assessment models." Transportmetrica A: Transport Science 17.4 (2021): 739-765.

[28] Coroama, Loredana, and Adrian Groza. "Explainable Artificial Intelligence for Person Identification." 2021 IEEE 17th International Conference on Intelligent Computer Communication and Processing (ICCP). IEEE, 2021.

[29] Gunning, David, et al. "XAI—Explainable artificial intelligence." Science robotics 4.37 (2019): eaay7120.

[30] Molnar, Christoph, Giuseppe Casalicchio, and Bernd Bischl. "Interpretable machine learning–a brief history, state-of-the-art and challenges." ECML PKDD 2020 Workshops: Workshops of the European Conference on Machine Learning and Knowledge Discovery in Databases (ECML PKDD 2020): SoGood 2020, PDFL 2020, MLCS 2020, NFMCP 2020, DINA 2020, EDML 2020, XKDD 2020 and INRA 2020, Ghent, Belgium, September 14–18, 2020, Proceedings. Cham: Springer International Publishing, 2021.

[31] Lin, Jianhua. "Divergence measures based on the Shannon entropy." IEEE Transactions on Information theory 37.1 (1991): 145-151.

[32] Klir, George J. "Uncertainty and information: foundations of generalized information theory." Kybernetes 35.7/8 (2006): 1297-1299.

[33] Klir, George, and Mark Wierman. Uncertainty-based information: elements of generalized information theory. Vol. 15. Springer Science \& Business Media, 1999.

[34] Altmann, André, et al. "Permutation importance: a corrected feature importance measure." Bioinformatics 26.10 (2010): 1340-1347.

[35] Gray, Robert M. Entropy and information theory. Springer Science \& Business Media, 2011.

[36] Bromiley, P. A., N. A. Thacker, and E. Bouhova-Thacker. "Shannon entropy, Renyi entropy, and information." Statistics and Inf. Series (2004-004) 9 (2004): 2-8.

[37] Cai, Yuhang, and Lek-Heng Lim. "Distances between probability distributions of different dimensions." IEEE Transactions on Information Theory 68.6 (2022): 4020-4031.

[38] Danielsson, Per-Erik. "Euclidean distance mapping." Computer Graphics and image processing 14.3 (1980): 227-248.

[39] Adadi, Amina, and Mohammed Berrada. "Peeking inside the black-box: a survey on explainable artificial intelligence (XAI)." IEEE access 6 (2018): 52138-52160.

[40] Rényi, Alfréd. "On measures of entropy and information." Proceedings of the Fourth Berkeley Symposium on Mathematical Statistics and Probability, Volume 1: Contributions to the Theory of Statistics. Vol. 4. University of California Press, 1961.

[41] Joyce, James M. "Kullback-leibler divergence." International encyclopedia of statistical science. Springer, Berlin, Heidelberg, 2011. 720-722.

[42] Menéndez, M. L., et al. "The jensen-shannon divergence." Journal of the Franklin Institute 334.2 (1997): 307-318.

[43] Baram, Yoram, Ran El Yaniv, and Kobi Luz. "Online choice of active learning algorithms." Journal of Machine Learning Research 5.Mar (2004): 255-291.

[44] Batina, Lejla, et al. "Mutual information analysis: a comprehensive study." Journal of Cryptology 24.2 (2011): 269-291.

[45] Linge, Yanis, Cécile Dumas, and Sophie Lambert-Lacroix. "Maximal Information Coefficient Analysis." Cryptology ePrint Archive (2014).

[46] Gu, Xiangyuan, et al. "Conditional mutual information-based feature selection algorithm for maximal relevance minimal redundancy." Applied Intelligence 52.2 (2022): 1436-1447.

[47] Li, Ming, and Paul Vitányi. An introduction to Kolmogorov complexity and its applications. Vol. 3. New York: Springer, 2008.

[48] Hammer, Daniel, et al. "Inequalities for Shannon entropy and Kolmogorov complexity." Journal of Computer and System Sciences 60.2 (2000): 442-464.

[49] Uspensky, Vladimir A. "Complexity and entropy: an introduction to the theory of Kolmogorov complexity." Kolmogorov complexity and computational complexity (1992): 85-102.
\newpage
\section{Supplementary}
\subsection{Table for Notation Referances}
\begin{table}[h]
    \centering
    \begin{tabular}{|l|l|}
    \hline
        Notations & Descriptions \\
        \hline
        $||F||^{(k\times n)} $   & $(k\times n)$ dimensional feature space.\\
        \hline
        $k$& Number of features in both original and sample datasets.\\
        \hline
        $n$& Number of observations in each feature in the original database.\\
        \hline
        $n'$& Number of observations in each feature in the sample database. \\
        \hline
        $\vec{Y}= (y_1,..,y_{n})$ & Output vector in original space.\\
        \hline
        $\vec{Y'}= (y'_1,..,y'_{n'})$& Output vector in sample space.\\
        \hline
        $\mathcal{U}$& Unified superspace for original model.\\
        \hline
        $\mathcal{U}'$& Unified superspace for lightweight model.\\
        \hline
        $\mathcal{D}_i(\vec{f_i}); \ i=1:k$ & Distribution of the $i ^{th}$ feature.\\
        \hline
        $\mathcal{D}(\vec{Y})$& Distribution of output in original space.\\
        \hline
        $\mathcal{D}(\vec{Y'})$& Distribution of output in lightweight space.\\
        \hline 
        $\vec{\mathcal{F}} = (\mathcal{F}_1, \mathcal{F}_2, ...., \mathcal{F}_n)'$ & Input column vector.\\
        \hline
         $ D^2_\text{final}$& The final average distance from the features to output in original space.\\
         \hline
         $ D^{'2}_\text{final}$ &The final average distance from the features to output in lightweight space.\\
        \hline 
        $\mu$ & Lightweight model output probability measure.\\
        \hline
        $\delta$& Original model probability measure.\\
        \hline
        $d^{+}$ & Embedding distance.\\
        \hline 
         $d^{-}$& Projection distance.\\
         \hline 
         $\mathcal{L}(Y,Y')$& Loss between original and lightweight model outputs.\\
         \hline 
         $\mathcal{R}(d*)$& Risk generator.\\
         \hline 
         $\lambda$& Regularized parameter.\\
         \hline
         $\hat{f_i}$ & The mean of the feature $f_i ; i = 1:k$. \\
         \hline
         $\hat{y'}$ & the mean of the output vector $Y$.\\
         \hline
         $\hat{f_iy'}$ & and the joint weighted mean of feature $f_i$ ; $i = 1:k$ and $Y$.\\
         \hline 
         $ \eta_{f_i}$ & $\mbox{PCIR}$ for feature $f_i$.\\
         \hline 
         $\mathfrak{C}_{f_i}$ & $\mbox{MCIR}$- Mutual Correlation Impact Ratio for feature $f_i$.\\
         \hline 
         $\mathfrak{J}$& $\mbox{JMI}$- Join Mutual Impact. \\
         \hline
    \end{tabular}
    \label{tab:my_label}
\end{table}
\subsection{Proof of theorem 1}\label{thm1}
\textbf{Proof:}
In the following, we show that each local output can be expressed as a combination of the feature input.\\
\textbf{case 1 :} When, $j \leq m$;
\begin{equation}
\begin{aligned}
    \frac{dy'}{d\vec{f_j}}  = &\frac{d}{d\vec{f_j}}( \frac{\eta_{f_1} \vec{f_1} + \eta_{f_2} \vec{f_2} + ...... + \eta_{f_{m}} \vec{f_{m}}}{\eta_{f_{m+1}} \vec{f_{m+1}} + \eta_{f_{m+2}} \vec{f_{m+2}} + ....... + \eta_{f_{k}} \vec{f_{k} }}) \\ 
   = &[\frac{d}{d\vec{f_j}}( \eta_{f_1} \vec{f_1}) + \frac{d}{df_j}(\eta_{f_2} \vec{f_2}) + ....+ \frac{d}{df_j}(\eta_{f_j} \vec{f_j})+.... + \frac{d}{df_j}(\eta_{f_{m}} \vec{f_{m}})] \times\\ &\frac{d}{df_j}[\frac{1}{\eta_{f_{m+1}} \vec{f_{m+1,j}} + \eta_{f_{m+2}} \vec{f_{m+2,j}} + ....... + \eta_{f_{k}} \vec{f_{kj}}}]\\
    =&[\eta_{f_1} \frac{d\vec{f_1}}{d\vec{f_j}} + \eta_{f_2}\frac{d\vec{f_2}}{d\vec{f_j}} + ....+ \eta_{f_j}\frac{d\vec{f_j}}{d\vec{f_j}}+.... + \eta_{f_m}\frac{d\vec{f_m}}{d\vec{f_j}}] \times \\ &\frac{d}{d\vec{f_j}}[\frac{1}{\eta_{f_{m+1}} \vec{f_{m+1,j}} + \eta_{f_{m+2}} \vec{f_{m+2,j}} + ....... + \eta_{f_{k}} \vec{f_{kj}}}]\\
\end{aligned}
\end{equation}
 Since the features are independent of each other when we consider the impact of the feature $\vec{f_j}$ on the output $Y'$, we fix the values of features. Then we have;

\begin{align}
    \frac{dy'}{d\vec{f_j}}  = \eta_{f_j}\frac{d\vec{f_j}}{d\vec{f_j}} \times \frac{1}{\mathbb{K}} =  c_1 * \eta_{f_j}
\end{align}
Here, $c_1 = \frac{1}{\mathbb{K}} = \frac{1}{\eta_{f_{m+1}} \vec{f_{m+1,j}} + \eta_{f_{m+2}} \vec{f_{m+2,j}} + ....... + \eta_{f_{k}} \vec{f_{kj}}}$ is a constant that can be expressed as the combination of the rest of the features and their correlation ratio values which are fixed in our environment. \\
\textbf{case 2:}
when $j > m$;
\begin{equation}
\begin{aligned}
    \frac{dy'}{d\vec{f_j}}  = &\frac{d}{d\vec{f_j}}( \frac{\eta_{f_1} \vec{f_1} + \eta_{f_2} \vec{f_2} + ...... + \eta_{f_{m}} \vec{f_{m}}}{\eta_{f_{m+1}} \vec{f_{m+1}} + \eta_{f_{m+2}} \vec{f_{m+2}} + ....... + \eta_{f_{k}} \vec{f_{k}} }) \\ 
    =&[\frac{d}{d\vec{f_j}}( \eta_{f_1} \vec{f_1}) + \frac{d}{d\vec{f_j}}(\eta_{f_2} \vec{f_2}) + ........ + \frac{d}{d\vec{f_j}}(\eta_{f_{m}} \vec{f_{m}})] \times\\ &\frac{d}{d\vec{f_j}}[\frac{1}{\eta_{f_{m+1}} \vec{f_{m+1,j}} + \eta_{f_{m+2}} \vec{f_{m+2,j} }+ ....+\eta_{f_j} \vec{f_j}+... + \eta_{f_{k}} \vec{f_{kj}}}]\\
    &
    [{\eta_{f_{m+1}} \vec{f_{m+1,j}} + \eta_{f_{m+2}} \vec{f_{m+2,j}} + ....+\eta_{f_j} \vec{f_j}+... + \eta_{f_{k}} \vec{f_{kj}}}]\frac{d}{df_j}(\eta_{f_1} \vec{f_1} + \eta_{f_2} \vec{f_2} + ...... + \eta_{f_{m}} \vec{f_m})\\ =\\& - (\eta_{f_1} \vec{f_1} + \eta_{f_2} \vec{f_2} + ...... + \eta_{f_{m}} \vec{f_m})\frac{d}{df_j} [{\eta_{f_{m+1}} \vec{f_{m+1,j}} + \eta_{f_{m+2}} \vec{f_{m+2,j}} + ....+\eta_{f_j} \vec{f_j}+... + \eta_{f_{k}} \vec{f_{kj}}}] \times \\ & \frac{1}{[\frac{d}{df_j}[\eta_{f_{m+1}} \vec{f_{m+1,j}} + \eta_{f_{m+2}} \vec{f_{m+2,j}} + ....+\eta_{f_j} \vec{f_j}+... + \eta_{f_{k}} \vec{f_{kj}}]^2]} \\
    =&\frac{-k-1. \eta_{f_j}}{2. [\eta_{f_{m+1}} \vec{f_{m+1,j}} + \eta_{f_{m+2}} \vec{f_{m+2,j}} + ....+\eta_{f_j} \vec{f_j}+... + \eta_{f_{k}} \vec{f_{kj}}] \frac{d}{d\vec{f_j}}[\eta_{f_j}^2 \vec{f_j}^2 - 2\eta_{f_j}(\eta_{f_1}+\eta_{f_2}+...\eta_{f_k})]}\\
    =&\frac{-k_1. \eta_{f_j}}{2.\eta_{f_j} (\eta_{f_j}^2 -2 K_2)} \\ 
    =&\frac{k_1}{2 (2 K_2-\eta_{f_j}^2) }     \\
    =& \frac{c_2}{(2 K_2-\eta_{f_j}^2) } 
    \end{aligned}
\end{equation}
Equations (2)-(3) imply that the correlation ratio is able to capture and reflect the impact of a certain feature on the generated output. Mathematically when the feature has a direct positive relation with the output, it should belong to the numerator of the complex model. On the other hand, the feature belonging to the denominator of the model shows a constant inverse relation with the output. In the second case, we can say that the features are conversely correlated to the output.

\subsection{Derivation of equation \ref{cmmi}}\label{thm2}
 For the sake of simplicity, Let consider we have only three features $\vec{f_1}, \vec{f_2}, \vec{f_3}$ dependent on each other. If we want to find the \mbox{CMMI} between $\vec{Y}$ and $\vec{f_1}$ where $\vec{f_1}$ is dependent on $\vec{f_2}$ and $\vec{ f_3}$. We also assume that $\vec{f_1}$ is continuous. On the other hand, $\vec{f_2}, \vec{f_3}$, and $\vec{Y}$ are discrete in nature. Then using the concept of \mbox{MI} [46], the \mbox{CMMI} will be :
\begin{equation}
\begin{aligned}
     I(\vec{Y'}; \vec{f_1} |\vec{f_2}, \vec{f_3}) &=  \sum_{f_3 }\sum_{f_2}\int_{f_1} \sum_{Y}P(Y, f_1,f_2, f_3) \log_2[\frac{P(Y, f_1|f_2,f_3)}{P(Y|f_2,f_3) P(f_1|f_2,f_3 )} ]\\
     &=  \sum_{f_3 }\sum_{f_2}\int_{f_1} \sum_{Y}P(Y, f_1,f_2, f_3) \log_2[\frac{\frac{P(Y,f_1,f_2,f_3)}{P(f_2,f_3)}}{P(Y|f_2,f_3) P(f_1|f_2,f_3 )} ]
\end{aligned}
\end{equation}
By using Bayes' theorem it can be expressed as :
\begin{equation}
    \begin{aligned}
       I(\vec{Y'}; \vec{f_1} |\vec{f_2}, \vec{f_3}) &=  \sum_{f_3 }\sum_{f_2}\int_{f_1} \sum_{Y'}P(Y', f_1,f_2,f_3) \log_2[\frac{\frac{P(Y'|f_1,f_2,f_3) P(f_1|f_2,f_3)P(f_2|f_3) P(f_3)}{P(f_2|f_3) P(f_3)}}{P(Y|f_2,f_3) P(f_1|f_2,f_3 )} ] \\
       & =  \sum_{f_3 }\sum_{f_2}\int_{f_1} \sum_{Y'}P(Y, f_1,f_2,f_3) \log_2[\frac{P(Y'|f_1,f_2,f_3) }{P(Y|f_2,f_3) } ]
    \end{aligned}
\end{equation}
Similarly, using Bayes' theorem, we get:
\begin{equation}
    \begin{aligned}
       I(\vec{Y'}; \vec{f_2} |\vec{f_1}, \vec{f_3}) &= \sum_{f_3 }\sum_{f_2}\int_{f_1} \sum_{Y'}P(Y', f_1,f_2,f_3) \log_2[\frac{P(Y'|f_1,f_2,f_3) }{P(Y'|f_1,f_3) } ] \\
       I(\vec{Y'}; \vec{f_3} |\vec{f_1}, \vec{f_2}) &= \sum_{f_3 }\sum_{f_2}\int_{f_1} \sum_{Y'}P(Y', f_1,f_2,f_3) \log_2[\frac{P(Y'|f_1,f_2,f_3) }{P(Y'|f_1,f_2) } ]
    \end{aligned}
\end{equation}
we can also write in the following way: 
\begin{equation}
    \begin{aligned}
     I(\vec{Y'}; \vec{f_1} |\vec{f_2}, \vec{f_3}) &= \sum_{f_3 }\sum_{f_2}\int_{f_1} \sum_{Y'}P(Y', f_1,f_2,f_3) \log_2[\frac{P(Y'|\{f_j\}\epsilon||F||^{k \times n'}) }{P(Y|\{f_j\}\epsilon\{||F||^{k \times n'} -f_1\}) } ]\\
       I(\vec{Y'}; \vec{f_2} |\vec{f_1}, \vec{f_3}) &= \sum_{f_3 }\sum_{f_2}\int_{f_1} \sum_{Y'}P(Y', f_1,f_2,f_3) \log_2[\frac{P(Y'|\{f_j\}\epsilon||F||^{k \times n'}) }{P(Y'|\{f_j\}\epsilon\{||F||^{k \times n'} -f_2\}) } ] \\
        I(\vec{Y'}; \vec{f_3} |\vec{f_1}, \vec{f_2}) &= \sum_{f_3 }\sum_{f_2}\int_{f_1} \sum_{Y'}P(Y', f_1,f_2,f_3) \log_2[\frac{P(Y'|\{f_j\}\epsilon||F||^{k \times n'}) }{P(Y'|\{f_j\}\epsilon\{||F||^{k \times n'} -f_3\}) } ]
    \end{aligned}
\end{equation}
 In the second step, we take four features $\vec{f_1}, \vec{f_2}, \vec{f_3}, \vec{f_4}$ depending on each other. Again, we assume, $\vec{f_1}$ is continuous and $\vec{f_4},\vec{f_3},\vec{f_2}$ , and $\vec{Y'}$ are discrete in nature. Then $\mbox{CMMI}$ of $\vec{Y'}$ and $(\vec{f_1}|\vec{f_2}, \vec{f_3}, \vec{f_4)}$ will be :
 \begin{equation}
     \begin{aligned}
         I(\vec{Y'}; \vec{f_1} |\vec{f_2}, \vec{f_3}, \vec{f_4}) &=\sum_{f_4}  \sum_{f_3 }\sum_{f_2}\int_{f_1} \sum_{Y'}P(Y', f_1,f_2, f_3,f_4) \log_2[\frac{P(Y', f_1|f_2,f_3,f_4)}{P(Y'|f_2,f_3,f_4) P(f_1|f_2,f_3,f_4 )} ]\\
         & = \sum_{f_4}  \sum_{f_3 }\sum_{f_2}\int_{f_1} \sum_{Y'}P(Y', f_1,f_2, f_3,f_4) \log_2[\frac{\frac{P(Y',f_1,f_2,f_3,f_4)}{P(f_2,f_3,f_4)}}{P(Y'|f_2,f_3,f_4) P(f_1|f_2,f_3,f_4 )} ]\\
         &=  \sum_{f_4}  \sum_{f_3 }\sum_{f_2}\int_{f_1} \sum_{Y'}P(Y', f_1,f_2, f_3,f_4) \log_2[\frac{\frac{P(Y'|f_1,f_2,f_3,f_4)P(f_1|f_2,f_3,f_4)P(f_2|f_3,f_4)P(f_3|f_4)P(f_4)}{P(f_2|f_3,f_4)P(f_3|f_4) P(f_4)}}{P(Y'|f_2,f_3,f_4) P(f_1|f_2,f_3,f_4 )} ] \\
         &= \sum_{f_4}  \sum_{f_3 }\sum_{f_2}\int_{f_1} \sum_{Y'}P(Y', f_1,f_2, f_3,f_4) \log_2[\frac{P(Y'|f_1,f_2,f_3,f_4)}{P(Y'|f_2,f_3,f_4)}]\\
         &= \sum_{f_4}  \sum_{f_3 }\sum_{f_2}\int_{f_1} \sum_{Y'}P(Y', f_1,f_2, f_3,f_4) \log_2[\frac{P(Y'|\{f_j\} \epsilon ||F||^{k \times n'})}{P(Y'|\{f_j\} \epsilon \{||F||^{k \times n'}-f_1\})}]
     \end{aligned}
 \end{equation}
 Similarly, it can be written as:
\begin{equation}
    \begin{aligned}
        I(\vec{Y'}; \vec{f_2} |\vec{f_1}, \vec{f_3}, \vec{f_4})&= \sum_{f_4}  \sum_{f_3 }\sum_{f_2}\int_{f_1} \sum_{Y'}P(Y', f_1,f_2, f_3,f_4) \log_2[\frac{P(Y'|\{f_j\} \epsilon ||F||^{k \times n'})}{P(Y'|\{f_j\} \epsilon \{||F||^{k \times n'}-f_2\})}]  \\
        I(\vec{Y'}; \vec{f_3} |\vec{f_2}, \vec{f_1}, \vec{f_4}) &= \sum_{f_4}  \sum_{f_3 }\sum_{f_2}\int_{f_1} \sum_{Y'}P(Y', f_1,f_2, f_3,f_4) \log_2[\frac{P(Y'|\{f_j\} \epsilon ||F||^{k \times n'})}{P(Y'|\{f_j\} \epsilon \{||F||^{k \times n'}-f_3\})}] \\
        I(\vec{Y'}; \vec{f_4} |\vec{f_2}, \vec{f_3}, \vec{f_1})&= \sum_{f_4}  \sum_{f_3 }\sum_{f_2}\int_{f_1} \sum_{Y'}P(Y', f_1,f_2, f_3,f_4) \log_2[\frac{P(Y'|\{f_j\} \epsilon ||F||^{k \times n'})}{P(Y'|\{f_j\} \epsilon \{||F||^{k \times n'}-f_4\})}] \\
    \end{aligned}
\end{equation}
In this way we can claim that: 
\begin{equation}
    \begin{aligned}
         &I(\vec{Y'};\vec{f_{i-1}} | \{\vec{f_j}\} \subseteq \{||F||^{k \times n' } - \vec{f_{i-1}}\}; i\neq j ))=  \\ & \sum_{f_j \subseteq \{||F||^{k \times n'} - f_{i-1}\}_d} \int_{f_j \subseteq \{||F||^{k \times n'} - f_{i-1}\}_c}\int_{f_{i-1}} \sum_{Y'}P(Y', f_1,f_2, .., f_i, f_j, ..f_k) \log_2[\frac{P(Y'| \{f_j\} \subseteq ||F||^{k \times n'} )}{P(Y'|\{f_j\} \subseteq \{||F||^{k \times n'} - f_{i-1}\}; i\neq j) } ]
    \end{aligned}
\end{equation}
So, if we consider the environment for $k$ number of features, where we want to find the $\mbox{CMMI}$ between $\vec{Y'}$ and $\vec{f_i}$; $i=1(1)k$. $\vec{f_i}$ is dependent on the set of the features $\{\vec{f_j}\} \epsilon \{||F||^{k \times n'} -f_i\} $ , 
\begin{equation}
    \begin{aligned}
        &I(\vec{Y'};\vec{f_i} | \{\vec{f_j}\} \subseteq \{||F||^{k \times n' } - \vec{f_i}\}; i\neq j )) = \\ & 
        \sum_{f_j \subseteq \mathcal{A}} \int_{f_j \subseteq \mathcal{B}}\int_{f_i} \sum_{Y'}\Phi \log_2[\frac{P(Y', f_i| \{f_j\} \subseteq \{||F||^{k \times n'} - f_i\}; i\neq j)}{P(Y'|\{f_j\} \subseteq \{||F||^{k \times n'} - f_i\}) P(f_i|\{f_j\} \subseteq \{||F||^{k \times n'} - f_i\} )} ] \\
        & \sum_{f_j \subseteq \mathcal{A}} \int_{f_j \subseteq \mathcal{B}}\int_{f_i} \sum_{Y'}\Phi  \log_2[\frac{\frac{P(Y', f_1, f_2, f_3, .., f_{i-1}, f_i, f_{i+1}, .., f_k)}{P(f_1,f_2,..f_{i-1},f_{i+1}, ..,f_k)}}{P(Y'|\{f_j\} \subseteq \{||F||^{k \times n'} - f_i\}) P(f_i|\{f_j\} \subseteq \{||F||^{k \times n'} - f_i\} )} ]\\
        & = \sum_{f_j \subseteq \mathcal{A}} \int_{f_j \subseteq \mathcal{B}}\int_{f_i} \sum_{Y'}\\&\Phi  \log_2[\frac{\frac{P(Y'| ||F||^{k \times n'}) P(f_i|\{f_j\}\epsilon\{||F||^{k \times n'}-f_i\}) P(f_1|\{f_j\}\epsilon\{||F||^{k \times n'}-f_i-f_1\}) P(f_2|\{f_j\}\epsilon\{||F||^{k \times n'}-f_i-f_1-f_2\})..P(f_k)}{  P(f_1|\{f_j\}\epsilon\{||F||^{k \times n'}-f_i-f_1\}) P(f_2|\{f_j\}\epsilon\{||F||^{k \times n'}-f_i-f_1-f_2\})..P(f_k)}}{P(Y'|\{f_j\} \subseteq \{||F||^{k \times n'} - f_i\}) P(f_i|\{f_j\} \subseteq \{||F||^{k \times n'} - f_i\} )} ]\\
        & = \sum_{f_j \subseteq \mathcal{A}} \int_{f_j \subseteq \mathcal{B}}\int_{f_i} \sum_{Y'}\Phi \log_2[\frac{P(Y'| ||F||^{k \times n'}) P(f_i|\{f_j\}\epsilon\{||F||^{k \times n'}-f_i\})}{P(Y'|\{f_j\} \subseteq \{||F||^{k \times n'} - f_i\}) P(f_i|\{f_j\} \subseteq \{||F||^{k \times n'} - f_i\} )} ]\\
        &= \sum_{f_j \subseteq \mathcal{A}} \int_{f_j \subseteq \mathcal{B}}\int_{f_i} \sum_{Y'}\Phi  \log_2[\frac{P(Y'| ||F||^{k \times n'}) }{P(Y'|\{f_j\} \subseteq \{||F||^{k \times n'} - f_i\})} ]\\
    \end{aligned}
\end{equation}
Here, $\mathcal{A} = \{||F||^{k \times n'}- f_i\}_d $ is the set that contains the discrete features and $\mathcal{B} = \{||F||^{k \times n'} - f_i\}_c$ is the set that contains continuous features. $\Phi = P(Y', f_1,f_2, .., f_i, f_j, ..f_k)$. \\
Therefore we can write, 
\begin{equation}
\begin{aligned}
 &I(\vec{Y'};\vec{f_i} | \{\vec{f_j}\} \subseteq \{||F||^{k \times n' } - \vec{f_i}\}; i\neq j )) = \\ & 
     \sum_{f_j \subseteq \{||F||^{k \times n'} - f_i\}_d} \int_{f_j \subseteq \{||F||^{k \times n'}- f_i\}_c}\int_{f_i} \sum_{Y'}P(Y', f_1,f_2, .., f_i, f_j, ..f_k)  \log_2[\frac{P(Y'| \{f_j\} \subseteq ||F||^{k \times n'} )}{P(Y'|\{f_j\} \subseteq \{||F||^{k \times n'}- f_i\}; i\neq j) } ]
    \end{aligned}
\end{equation}
$\mbox{CMMI}$ could take the values between $0$ to $\infty$, unlike the strict bound of $(0,1)$ that the normal correlation coefficient has. The infinite range can be the cause of the problem regarding scalability. So, it will be better to scale down the dependency. This problem can be solved by our proposed metrics Mutual Correlation Impact Ratio ($\mbox{MCIR}$) ratio. $\mbox{MCIR}$ is the ratio of two mutual information which is defined below. $\mbox{MCIR}$ has a strict bound between $0$ to $1$.   \\
\subsection{proof of result 1}\label{rslt1}
\textbf{Proof: }
The mutual correlation impact ratio is a ratio that measures the relation distance between a target feature and the output variable given the fact that the target feature is dependent on some other features. More specifically, $\mbox{MCIR}$ measures how much influence other features have in the dependence of the target feature and the output. The following Venn diagram will help to understand how the feature-wise $\mbox{MCIR}$ is calculated. (upload later)
Now by the definition of MI [44], it is known that the value of Mutual information varies between $0$ to $\infty$. So we can say that,
\begin{equation}
    0 \leq I(\vec{Y'};\vec{f_i}|{\vec{f_j}}\subseteq {||F||^{k \times n'} -\vec{f_i}}; i\neq j) \leq \infty
\end{equation}
Now, obviously, 
\begin{equation}
    0 \leq I(\vec{Y'};\vec{f_i}|{\vec{f_j}}\subseteq {||F||^{k \times n'} -\vec{f_i}}; i\neq j) \leq \leq I(\vec{Y'};\vec{f_i}|{\vec{f_j}}\subseteq {||F||^{k \times n'} -\vec{f_i}}; i\neq j) + I(\vec{Y'}, \vec{f_1}, \vec{f_2}, ..., \vec{f_k}) \leq \infty
\end{equation}
as, $I(\vec{Y'},\vec{f_1},\vec{f_2},\vec{f_3},...., \vec{f_k})\geq 0$; so it is very obvious that,
\begin{equation}
    0 \leq \frac{I(\vec{Y'};\vec{f_i} | \{\vec{f_j}\} \subseteq \{||F||^{k \times n'} - \vec{f_i}\}; i\neq j )}{I(\vec{Y'};\vec{f_i} | \{\vec{f_j}\} \subseteq \{||F||^{k \times n'} - \vec{f_i}\}; i\neq j )+ I(\vec{Y'}, \vec{f_1}, \vec{f_2}, ...\vec{f_{j-1}}, \vec{f_j}, \vec{f_{j+1}}, ..,\vec{ f_k)}} \leq 1
\end{equation}
\subsection{ Complexity bound on MCIR}\label{complex}
Feature dependencies create a complex environment where computational complexity can not be ignored. We have used the concept of Mutual Information to calculate our desired metrics MCIR. In this section, we will calculate the upper bound of the computation complexity when the features are dependent on each other. We will use the notion of Kolmogorov Complexity [47] to compute the complexity bound. Let assume our model for feature dependency is $M(f_{j(f,\mathfrak{C})}) = M*$ and the universal computer is $\mathbb{C}$. by the definition of Kolmogorov Complexity[48] we can write, 
\begin{equation}
    K_\mathbb{C}(\vec{Y'}| I(\vec{Y'})) = \min_{M*: \mathbb{C}(M*, I(\vec{Y'})) = \vec{Y'}} I(M*)
\end{equation}
Here, $I(Y')$ denotes the length of the output string $\vec{Y'}$, $I(M*)$ denotes the length of the string of the program of the model $M*$, and  $M*$ is the model running on the universal computer $\mathbb{C}$ to produce the output $y'$.
Then, by the theorem of Universality of conditional Kolmogorov complexity [49] knowing $I(\vec{Y'})$, we can find the upper bound: 
\begin{equation}
    K_\mathbb{C}(\vec{Y'}) \leq  K_\mathbb{A}(\vec{Y'}| I(\vec{Y'})) + \log^*I(\vec{Y'})+ Const_\mathbb{A}
\end{equation}
Where, $\log^* I(\vec{Y'})=  \log I(\vec{Y'}) + \log\log I(\vec{Y'}) +\log\log\log I(\vec{Y'})+......$ as long as the terms are positive. $k$ is the total number of features in the data.  $\mathbb{A}$ is any other computer and there exist a constant $Const_\mathbb{A}$ for all strings $\vec{Y'} = (y'_1, y'_2, ..., y'_{n'}) \epsilon \{0,1\}$. $Const_\mathbb{A}$ does not depends on the output $\vec{Y'}$. Here, $I(\vec{Y'})= n'$, then, it can be written as,
\begin{equation}\label{kol}
    K_\mathbb{C}(\vec{Y'}) \leq  K_\mathbb{A}(\vec{Y'}| I(\vec{Y'})) + \frac{k}{2}O(\log{n'})+ Const_\mathbb{A}
\end{equation}
where we take $O(\log{(\log{(n')})}) = O(\log{n'})$ and ignored the negetive terms. 
We will also calculate the complexity bound of each $I(\vec{Y'};\vec{f_i} | \{\vec{f_j}\} \subseteq \{||F||^{k \times n'} - \vec{f_i}\}; i\neq j );  i=1:k$ and by the theory of Kolmogorov Complexity bound for entropy [47], it should be as follows:
\begin{equation}\label{big}
    \begin{aligned}
       & I(\vec{Y'};\vec{f_i} | \{\vec{f_j}\} \subseteq \{||F||^{k \times n'} - \vec{f_i}\}; i\neq j ) = H(\vec{Y'}, \vec{f_j})- H(\vec{Y'};\vec{f_i} | \{\vec{f_j}\} \subseteq \{||F||^{k \times n'} - \vec{f_i}\}; i\neq j)\\
        & \leq H(\vec{Y})- H(\vec{Y'};\vec{f_i} | \{\vec{f_j}\} \subseteq \{||F||^{k \times n'} - \vec{f_i}) + \frac{(||Y||^{k*n'} \log{n'})}{n'} - \frac{([||Y||-1]^{k*n'} \log{n'})}{n'}+ \frac{Const_\mathbb{A}}{n'}
    \end{aligned}
\end{equation} 
By taking $ \frac{(||Y||^{k*n'} \log{n'})}{n'} - \frac{([||Y||-1]^{k*n'} \log{n'})}{n'} =  O(\log{n'}) $
we can also rewrite the equation \ref{big} following way:
\begin{equation} \label{a1}
    \begin{aligned}
       I(\vec{Y'};\vec{f_i} | \{\vec{f_j}\} \subseteq \{||F||^{k \times n'} - \vec{f_i}\}; i\neq j )& \leq I(\vec{Y'};\vec{f_i} | \{\vec{f_j}\} \subseteq \{||F||^{k \times n'} - \vec{f_i}\}; i\neq j ) + O(\log{n'}) + \frac{Const_\mathbb{A}}{n'}\\
        I(\vec{Y'};\vec{f_i} | \{\vec{f_j}\} \subseteq \{||F||^{k \times n'} - \vec{f_i}\}; i\neq j ) &\leq  O(\log{n'}) + \mathbb{C_A}_i
    \end{aligned}
\end{equation}
where, $\mathbb{C_A}_i = I(\vec{Y'};\vec{f_i} | \{\vec{f_j}\} \subseteq \{||F||^{k \times n'} - \vec{f_i}\}; i\neq j )  + \frac{Const_\mathbb{A}}{n'} $
Similarly, by Kolmogorov's theorem [47][49], we also can get, 
\begin{equation}
    I(\vec{Y'}, \vec{f_1}, \vec{f_2}, ...\vec{f_{j-1}}, \vec{f_j}, \vec{f_{j+1}}, ..,\vec{ f_k}) \leq  I(\vec{Y'}, \vec{f_1}, \vec{f_2}, ...\vec{f_{j-1}}, \vec{f_j}, \vec{f_{j+1}}, ..,\vec{ f_k}) + O(\log{n'}) +\frac{Const_\mathbb{A}}{n'}
    \end{equation}
Let denote, $ \frac{Const_\mathbb{A}}{n'} + I(\vec{Y'}, \vec{f_1}, \vec{f_2}, ...\vec{f_{j-1}}, \vec{f_j}, \vec{f_{j+1}}, ..,\vec{ f_k}) = \mathbb{C'_A}_i $ and  we will get the following in-equation for $i = 1:k$:
\begin{equation} \label{a2}
    I(\vec{Y'}, \vec{f_1}, \vec{f_2}, ...\vec{f_{j-1}}, \vec{f_j}, \vec{f_{j+1}}, ..,\vec{ f_k}) \leq   O(\log{n'}) + \mathbb{C'_A}_i 
    \end{equation}
Now, from equation \ref{a1} and equation \ref{a2}, we can get the complexity bound for calculating each $\mbox{MCIR}$, and it will be as follows for $i = 1:k$;
\begin{equation}\label{a3}
    \mathfrak{C}_i \leq \frac{1}{2}O(\log{n'}) + \zeta_\mathbb{A}
\end{equation}
here, $\zeta_\mathbb{A}$ is a constant associated with $ith$ $\mbox{MCIR}$ for any other computer $\mathbb{A}$; $i= 1, k$.
So, we can say from the equation \ref{kol} and equation \ref{a3} we can say that, our model would take no more than $\frac{k}{2}O(\log{n})+ Const_\mathbb{A}$ time complexity to run the whole program, whereas it would take no more than $\frac{1}{2}O(\log{n'}) + \zeta_\mathbb{A}$ time complexity to calculate each $\mbox{MCIR}$ associated with features.
\end{document}